\documentclass{article} 
\usepackage{nips14submit_e,times}
\usepackage{hyperref}
\usepackage{url}
\usepackage[utf8]{inputenc}
\usepackage{stmaryrd}
\usepackage{graphicx}
\usepackage{amsmath}

\nipsfinalcopy 

\newcommand{\bW}{{\mathbf{W}}}

\newcommand{\bu}{{\mathbf{u}}}
\newcommand{\bx}{{\mathbf{x}}}
\newcommand{\by}{{\mathbf{y}}}

\newcommand{\ignore}[1]{}

\title{Learning Multi-Relational Semantics Using Neural-Embedding Models}

\author{
Bishan Yang\thanks{Work conducted while interning at Microsoft Research.} \\
Cornell University \\
  Ithaca, NY, 14850, USA \\
  \texttt{bishan@cs.cornell.edu} \\
\And
Wen-tau Yih, Xiaodong He, Jianfeng Gao, Li Deng\\
Microsoft Research \\
  Redmond, WA 98052, USA \\
\texttt{scottyih,xiaohe,jfgao,deng@microsoft.com}
}

\begin{document}

\maketitle
Real-world entities (e.g.,~people and places) are often connected via relations, forming multi-relational data. Modeling multi-relational data is important in many research areas, from natural language processing to biological data mining~\cite{domingos2003prospects}. Prior work on multi-relational learning can be categorized into three categories: (1) statistical relational learning (SRL)~\cite{GetoorTa07}, such as Markov-logic networks~\cite{richardson2006markov}, which directly encode multi-relational graphs using probabilistic models; (2) path ranking methods~\cite{lao2011random,dong2014knowledge}, which explicitly explore the large relational feature space of relations with random walk; and (3) embedding-based models, which embed multi-relational knowledge into low-dimensional representations of entities and relations via tensor/matrix factorization~\cite{singh2008relational,NickelTrKr11,nickel2012factorizing}, Bayesian clustering framework~\cite{kemp2006learning, sutskever2009modelling}, and neural networks~\cite{paccanaro2001learning,bordes2013energy,BordesUsGaWeYa2013,SocherChenManningNg2013}. Our work focuses on the study of neural-embedding models, where the representations are learned in a neural network architecture. They have shown to be powerful tools for multi-relational learning and inference due to their high scalability and strong generalization abilities.

A number of techniques have been recently proposed to learn entity and relation representations using neural networks~\cite{bordes2013energy,BordesUsGaWeYa2013,SocherChenManningNg2013}. They all represent entities as low-dimensional vectors and represent relations as operators that combine the representations of two entities. The main difference among these techniques lies in the parametrization of the relation operators. For instance, given two entity vectors, the model of Neural Tensor Network (NTN)~\cite{SocherChenManningNg2013} represents each relation as a bilinear tensor operator and a linear matrix operator. The model of TransE~\cite{BordesUsGaWeYa2013}, on the other hand, represents each relation as a single vector that linearly interacts with the entity vectors. Both models report promising performance on predicting unseen relationships in knowledge bases. However, they have not been directly compared in terms of the different choices of relation operators and of the resulting effectiveness. Neither has the design of entity representations in these recent studies been carefully explored. For example, NTN~\cite{SocherChenManningNg2013} first shows the benefits of representing entities as an average of word vectors and initializing word vectors with pre-trained vectors from large text corpora. This idea is promising as pre-trained vectors tend to capture syntactic and semantic information from natural language and can assist in better generalization of entity embeddings. However, many real-world entities are expressed as non-compositional phrases (e.g. person names, movie names, etc.), of which meaning cannot be composed by their constituent words. Therefore, averaging word vectors may not provide an appropriate representation for such entities.

In this paper, we examine and compare different types of relation operators and entity vector representations under a general framework for multi-relational learning. Specifically, we derive several recently proposed embedding models, including TransE~\cite{BordesUsGaWeYa2013} and NTN~\cite{SocherChenManningNg2013}, and their variants under the same framework. We empirically evaluate their performance on a knowledge base completion task using various real-world datasets in a controlled experimental setting and present several interesting findings.
First, the models with fewer parameters tend to be better than more complex models in terms of both performance and scalability. Second, the bilinear operator plays an important role in capturing entity interactions. Third, with the same model complexity, multiplicative operations are superior to additive operations in modeling relations. Finally, initializing entity vectors with pre-trained phrase vectors can significantly boost performance, whereas representing entity vectors as an average of word vectors that are initialized with pre-trained vectors may hurt performance. These findings have further inspired us to design a simple knowledge base embedding model that significantly outperforms existing models in predicting unseen relationships, with a top-10 accuracy of 73.2\% (vs. 54.7\% by TransE) evaluated on Freebase.

\section{A General Framework for Multi-Relational Representation Learning}
\label{sec:framework}
Most existing neural embedding models for multi-relational learning can be derived from a general framework. The input is a relation triplet $(e_1,r,e_2)$ describing $e_1$ (the~\textit{subject}) and $e_2$ (the~\textit{object}) that are in a certain relation $r$. The output is a scalar measuring the validity of the relationship. Each input entity can be represented as a high-dimensional sparse vector (``one-hot" index vector or ``$n$-hot" feature vector). The first neural network layer projects the input vectors to low dimensional vectors, and the second layer projects these vectors to a real value for comparison via a relation-specific operator (it can also be viewed as a scoring function). 

More formally, denote $\bx_{e_i}$ as the input for entity $e_i$ and $\bW$ as the first layer neural network parameter. The scoring function for a relation triplet $(e_1, r, e_2)$ can be written as
\begin{equation}
\label{framework}
S_{(e_1, r, e_2)}=G_r\big(\by_{e_1}, \by_{e_2}\big),\;\mathrm{where} \;\by_{e_1}=f\big(\bW \bx_{e_1}\big),~~ \by_{e_2}=f\big(\bW \bx_{e_2}\big)
\end{equation}
Many choices for the form of the scoring function $G_r$ are available. Most of the existing scoring functions in the literature can be unified based on a basic linear transformation $g_r^a$, a bilinear transformation $g_r^b$ or their combination, where $g_r^a$ and $g_r^b$ are defined as
\begin{equation}
g_r^a(\by_{e_1},\by_{e_2})=\mathbf{A}_r^T\left(\begin{array}{c}\by_{e_1}\\ \by_{e_2}\end{array}\right)~~\textrm{and} ~~~g_r^b(\by_{e_1},\by_{e_2})=\by_{e_1}^T \mathbf{B}_r \by_{e_2},
\label{g}
\end{equation}
which are parametrized by $\mathbf{A}_r$ and $\mathbf{B}_r$, respectively.

In Table~\ref{summary}, we summarize several popular scoring functions in the literature for a relation triplet~$(e_1,r,e_2)$, reformulated in terms of the above two functions. Denote by $\by_{e_1},\by_{e_2}\in \it{R}^n$ two entity vectors. 
Denote by $\mathbf{Q}_{r_1}, \mathbf{Q}_{r_2}\in \it{R}^{m\times n}$ and $\mathbf{V}_r\in \it{R}^n$ matrix or vector parameters for linear transformation $g_r^a$ . 
Denote by   $\mathbf{M}_r\in \it{R}^{n \times n}$ and $\mathbf{T}_r\in \it{R}^{n\times n\times m}$ matrix or tensor parameters for bilinear transformation $g_r^b$.
$\mathbf{I}\in \it{R}^n$ is an identity matrix. $\bu_r \in \it{R}^{m}$ is an additional parameter for relation $r$. The scoring function for TransE is derived from $||\by_{e_1}-\by_{e_2}+V_r||_2^2$ as in~\cite{BordesUsGaWeYa2013}.

\begin{table*}[bth]
\begin{center}
\begin{footnotesize}
\begin{tabular}{|c|c|c|c|}
\hline
Models & $\mathbf{B}_r$ & $\mathbf{A}_r$ & Scoring Function $G_r$\\
\hline
Distance~\cite{bordes2011learning} & - & $\big[\mathbf{Q}_{r_1}\;~\;-\mathbf{Q}_{r_2}\big]$ & $-||g_r^a(\by_{e_1}, \by_{e_2})||_1$\\
\hline
Single Layer~\cite{SocherChenManningNg2013} & - & $\big[\mathbf{Q}_{r1}\;~~~~\;\mathbf{Q}_{r2}\big]$ & $\bu_r^T \tanh(g_r^a(\by_{e_1},\by_{e_2}))$\\
\hline
TransE~\cite{BordesUsGaWeYa2013}& $\mathbf{I}$ & $\big[\mathbf{V}_r^T\;~\;-\mathbf{V}_r^T\big]$ & $2g_r^a(\by_{e_1}, \by_{e_2})-2g_r^b(\by_{e_1}, \by_{e_2})+||\mathbf{V}_r||_2^2$\\
\hline
Bilinear~\cite{jenatton2012latent} & $\mathbf{M}_r$ & - & $g_r^b(\by_{e_1}, \by_{e_2})$\\
\hline
NTN~\cite{SocherChenManningNg2013} & $\mathbf{T}_r$ & $\big[\mathbf{Q}_{r1}\;\;~~~~\mathbf{Q}_{r2}\big]$ & $\bu_r^T \tanh\big(g_r^a(\by_{e_1},\by_{e_2})+g_r^b(\by_{e_1},\by_{e_2})\big)$\\
\hline
\end{tabular}
\end{footnotesize}
\caption{\label{summary} Comparisons among several multi-relational models in their scoring functions. }
\end{center}
\end{table*}

This general framework for relationship modeling also applies to the recent deep-structured semantic model~\cite{Huang-2013,Shen-2014,shen2014learning,Gao-2014,yihsemantic}, which learns the relevance or a single relation between a pair of word sequences. The framework above applies when using multiple neural network layers to project entities and using a relation-independent scoring function $G_r\big(\by_{e_1}, \by_{e_2}\big)=\cos[\by_{e_1}(\mathbf{W}_r),\by_{e_2}(\mathbf{W}_r)]$. The cosine scoring function is a special case of $g_r^b$ with normalized $\by_{e_1},\by_{e_2}$ and $\mathbf{B}_r=\mathbf{I}$.

The neural network parameters of all the models discussed above can be learned by minimizing a margin-based ranking objective\footnote{Other objectives such as mutual information (as in~\cite{Huang-2013}) and reconstruction loss (as in tensor decomposition approaches~\cite{CYYM14}) can also be applied. Comparisons among these objectives are beyond the scope of this paper.}, which encourages the scores of positive relationships (triplets) to be higher than the scores of any negative relationships (triplets). Usually only positive triplets are observed in the data. Given a set of positive triplets $T$, we can construct a set of negative triplets $T'$ by corrupting either one of the relation arguments, $T'=\{(e_1',r,e_2)|e_1'\in E,(e_1',r,e_2)\notin T\}\cup \{(e_1,r,e_2')|e_2'\in E, (e_1,r,e_2')\notin T\}$. The training objective is to minimize the margin-based ranking loss
\begin{equation}
\label{obj}
L(\Omega)=\sum_{(e_1,r,e_2)\in T}\sum_{(e_1',r,e_2')\in T'}\max\{S_{(e_1',r,e_2')}-S_{(e_1,r,e_2)}+1, 0\}
\end{equation}
\section{Experiments and Discussion}
\label{sec:comparison}

\paragraph{Datasets and evaluation metrics} We used the WordNet (WN) and Freebase (FB15k) datasets introduced in~\cite{BordesUsGaWeYa2013}. WN contains $151,442$ triplets with $40,943$ entities and $18$ relations, and FB15k consists of $592,213$ triplets with $14,951$ entities and $1345$ relations. We also consider a subset of FB15k (FB15k-401) containing only frequent relations (relations with at least $100$ training examples). This results in $560,209$ triplets with $14,541$ entities and $401$ relations.  
We use link prediction as our prediction task as in~\cite{BordesUsGaWeYa2013}. For each test triplet, each entity is treated as the target entity to be predicted in turn. Scores are computed for the correct entity and all the corrupted entities in the dictionary and are ranked in descending order. We consider \textit{Mean Reciprocal Rank (MRR)} (an average of the reciprocal rank of an answered entity over all test triplets),
\textit{HITS@10} (top-10 accuracy), and \textit{Mean Average Precision} (MAP) (as used in~\cite{CYYM14}) as the evaluation metrics.

\paragraph{Implementation details} All the models were implemented in C\# and using GPU. Training was implemented using mini-batch stochastic gradient descent with AdaGrad~\cite{duchi2011adaptive}. At each gradient step, we sampled for each positive triplet two negative triplets, one with a corrupted subject entity and one with a corrupted object entity. The entity vectors are renormalized to have unit length after each gradient step (it is an effective technique that empirically improved all the models). For the relation parameters, we used standard L2 regularization. For all models, we set the number of mini-batches to $10$, the dimensionality of the entity vector $d=100$, the regularization parameter $0.0001$, and the number of training epochs $T=100$ on FB15k and FB15k-401 and $T=300$ on~WN ($T$ was determined based on the learning curves where the performance of all models plateaued.) The learning rate was initially set to $0.1$ and then adapted during training by AdaGrad. 

\subsection{Model Comparisons}
We examine five embedding models in decreasing order of complexity: (1) NTN with $4$ tensor slices as in~\cite{SocherChenManningNg2013}; (2) Bilinear+Linear, NTN with $1$ tensor slice and without the non-linear layer; (3) TransE with L2 norm\footnote{Empirically we found no significant differences between L1-norm and L2-norm for the TransE objective.}
, a special case of Bilinear+Linear as described in~\cite{BordesUsGaWeYa2013}; (4) Bilinear; (5) Bilinear-diag: a special case of Bilinear where the relation matrix is a diagonal matrix.

\begin{table*}[bth]
\begin{center}
\begin{footnotesize}
\scalebox{0.9}{
\begin{tabular}{|c|c|c|c|c|c|c|}
\hline
& \multicolumn{2}{|c|}{FB15k}& \multicolumn{2}{|c|}{FB15k-401}& \multicolumn{2}{|c|}{WN}\\
& MRR & HITS$@$10 & MRR & HITS$@$10 & MRR & HITS$@$10\\
\hline
NTN & 0.25 & 41.4 & 0.24 & 40.5 & 0.53 & 66.1 \\
\hline
Blinear+Linear & 0.30 & 49.0 & 0.30 & 49.4 & 0.87 & 91.6\\
\hline
TransE ({\sc DistADD}) & 0.32 & 53.9 & 0.32 & 54.7 & 0.38 & 90.9\\
\hline
Bilinear & 0.31 & 51.9 & 0.32 & 52.2 & \textbf{0.89} & 92.8 \\
\hline
Bilinear-diag ({\sc DistMult}) & \textbf{0.35} & \textbf{57.7} & \textbf{0.36} & \textbf{58.5} & 0.83 & \textbf{94.2} \\
\hline
\end{tabular}
}
\end{footnotesize}
\caption{\label{comparison}Comparison of different embedding models}
\end{center}
\end{table*}

Table~\ref{comparison} shows the results of all compared methods on all the datasets. In general, we observe that the performance increases as the complexity of the model decreases on FB. NTN, the most complex model, provides the worst performance on both FB and WN, which suggests overfitting. Compared to the previously published results of TransE~\cite{BordesUsGaWeYa2013}, our implementation achieves much better results~(53.9\% vs. 47.1\% on FB15k and 90.9\% vs. 89.2\% on WN) using the same evaluation metric~(HITS@10). We attribute such discrepancy mainly to the different choice of SGD optimization: AdaGrad vs. constant learning rate. We also found that Bilinear consistently provides comparable or better performance than TransE, especially on WN. Note that WN contains much more entities than FB, it may require the parametrization of relations to be more expressive to better handle the richness of entities. Interestingly, we found that a simple variant of Bilinear -- {\sc Bilinear-diag}, clearly outperforms all baselines on FB and achieves comparable performance to Bilinear on WN.

\subsection{Multiplicative vs. Additive Interactions} 
Note that {\sc Bilinear-diag} and {\sc TransE} have the same number of model parameters and their difference lies in the operational choices of the composition of two entity vectors -- {\sc Bilinear-diag} uses weighted element-wise dot product (multiplicative operation) and {\sc TransE} uses element-wise subtraction with a bias (additive operation). To highlight the difference, here we use {\sc DistMult} and {\sc DistAdd} to refer to {\sc Bilinear-diag} and {\sc TransE}, respectively.
Comparison between these two models can provide us with more insights on the effect of two common choices of compositional operations -- multiplication and addition for modeling entity relations. Overall, we observed superior performance of {\sc DistMult} on all the datasets in Table~\ref{comparison}. 
Table~\ref{details} shows the \textit{HITS@10} score on four types of relation categories (as defined in~\cite{BordesUsGaWeYa2013}) on FB15k-401 when predicting the subject entity and the object entity respectively. We can see that {\sc DistMult} significantly outperforms {\sc DistAdd} in almost all the categories. More qualitative results can be found in the Appendix.


\begin{table*}[bth]
\begin{center}
\begin{footnotesize}
\scalebox{0.9}{
\begin{tabular}{|c|c|c|c|c|c|c|c|c|}
\hline
& \multicolumn{4}{|c|}{Predicting subject entities}& \multicolumn{4}{|c|}{Predicting object entities}\\
\hline
& 1-to-1 & 1-to-n & n-to-1 & n-to-n & 1-to-1 & 1-to-n & n-to-1 & n-to-n\\
\hline
{\sc DistADD} & 70.0 & 76.7 & 21.1 & 53.9 & 68.7 & 17.4 &~\textbf{83.2} & 57.5\\
\hline
{\sc DistMult} & \textbf{75.5} & \textbf{85.1} & \textbf{42.9} & \textbf{55.2} & \textbf{73.7} & \textbf{46.7} & 81.0 & \textbf{58.8}\\
\hline
\end{tabular}
}
\end{footnotesize}
\caption{\label{details}Results by relation categories: one-to-one, one-to-many, many-to-one and many-to-many}
\end{center}
\end{table*}
\subsection{Entity Representations}  
In the following, we examine the learning of entity representations and introduce two further improvements: using non-linear projection and initializing entity vectors with pre-trained phrase vectors. We focus on {\sc DistMult} as our baseline and compare it with the two modifications {\sc DistMult}-tanh (using $f=\tanh$ for entity projection\footnote{When applying non-linearity, we remove the normalization steps on entity parameters during training as $tanh$ already helps control the scaling freedoms.}) and {\sc DistMult}-tanh-EV-init (initializing the entity parameters with the $1000$-dimensional pre-trained phrase vectors released by \textit{word2vec}~\cite{mikolov2013distributed}) on FB15k-401. We also reimplemented the word vector representation and initialization technique introduced in~\cite{SocherChenManningNg2013} -- each entity is represented as an average of its word vectors and the word vectors are initialized using the $300$-dimensional pre-trained word vectors released by \textit{word2vec}. We denote this method as {\sc DistMult}-tanh-WV-init. Inspired by~\cite{CYYM14}, we design a new evaluation setting where the predicted entities are automatically filtered according to ``entity types" (entities that appear as the subjects/objects of a relation have the same type defined by that relation). This provides us with better understanding of the model performance when some entity type information is provided. 
\begin{table*}[bth]
\begin{center}
\scalebox{0.9}{
\begin{tabular}{|c|c|c|c|}
\hline
& MRR & HITS$@$10 & MAP (w/ type checking)\\
\hline
{\sc DistMult} & 0.36 & 58.5 & 64.5\\
\hline
{\sc DistMult}-tanh & 0.39 & 63.3 & 76.0\\
\hline
{\sc DistMult}-tanh-WV-init & 0.28 & 52.5 & 65.5\\
\hline
{\sc DistMult}-tanh-EV-init & \textbf{0.42} & \textbf{73.2} & \textbf{88.2}\\
\hline
\end{tabular}
}
\caption{\label{further_exp}Evaluation with pre-trained vectors}
\end{center}
\end{table*}

In Table~\ref{further_exp}, we can see that {\sc DistMult}-tanh-EV-init provides the best performance on all the metrics. Surprisingly, we observed performance drops by {\sc DistMult}-tanh-WV-init. We suspect that this is because word vectors are not appropriate for modeling entities described by non-compositional phrases (more than 73\% of the entities in FB15k-401 are person names, locations, organizations and films). The promising performance of {\sc DistMult}-tanh-EV-init suggests that the embedding model can greatly benefit from pre-trained entity-level vectors.

\section{Conclusion}
\label{sec:conclusion}
In this paper we present a unified framework for modeling multi-relational representations, scoring, and learning, and conduct an empirical study of several recent multi-relational embedding models under the framework. We investigate the different choices of relation operators based on linear and bilinear transformations, and also the effects of entity representations by incorporating unsupervised vectors pre-trained on extra textual resources. Our results show several interesting findings, enabling the design of a simple embedding model that achieves the new state-of-the-art performance on a popular knowledge base completion task evaluated on Freebase. Given the recent successes of deep learning in various applications; e.g. \cite{Hinton2012,Vinyals12,deng2013new}, our future work will aim to exploit deep structure including possibly tensor construct in computing the neural embedding vectors; e.g. \cite{Huang-2013,Yu2013,HutchinsonPAMI}. This will extend the current multi-relational neural embedding model to a deep version that is potentially capable of capturing hierarchical structure hidden in the input data.


\bibliographystyle{plain}
\bibliography{ref}

\section*{Appendix}
\begin{figure}[ht]
\centering
\includegraphics[width=.8\linewidth]{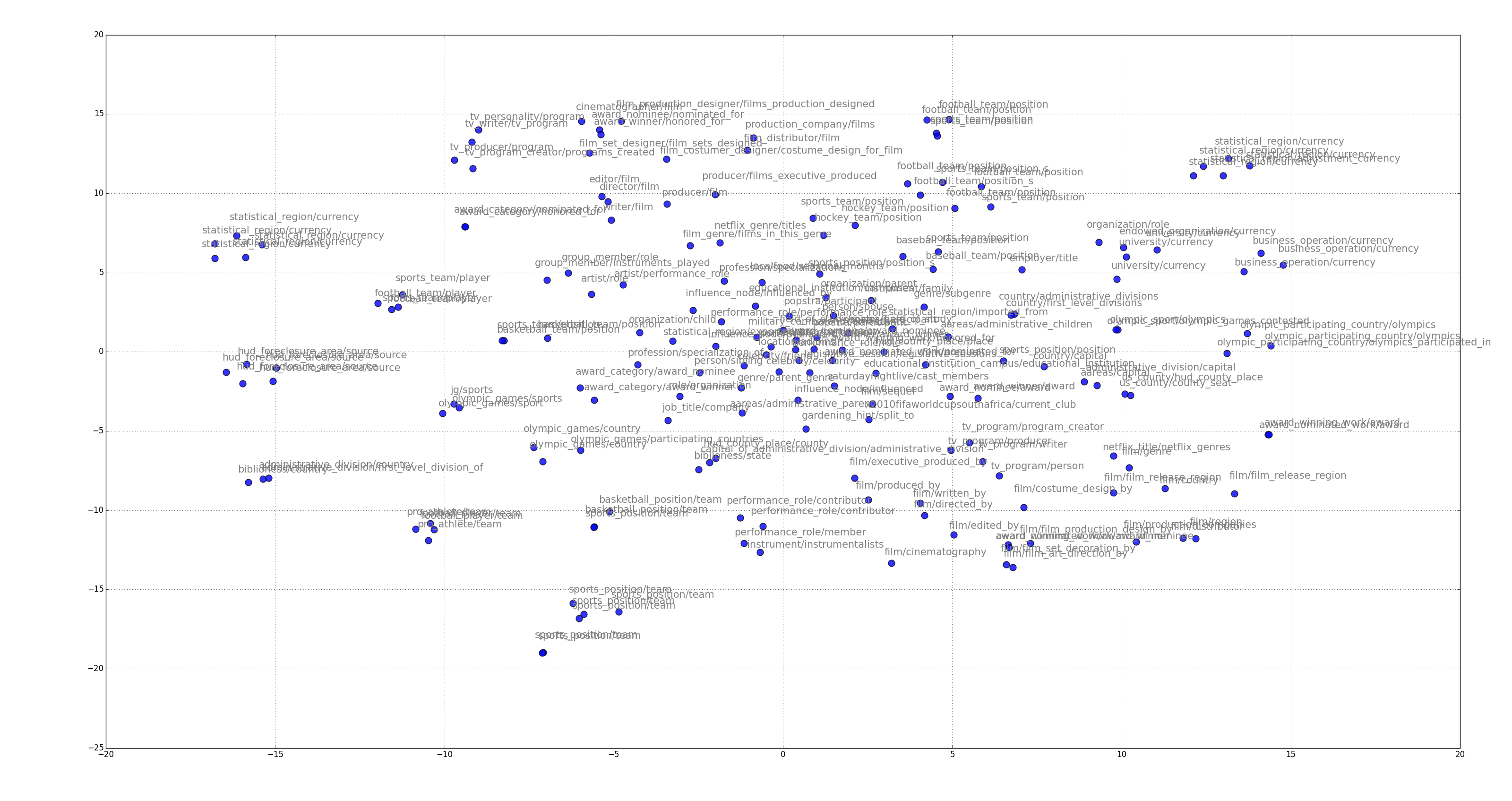}
\caption{Relation embeddings ({\sc DistAdd})}
\label{fig:distadd}
\end{figure}

\begin{figure}[ht]
\centering
\includegraphics[width=.8\linewidth]{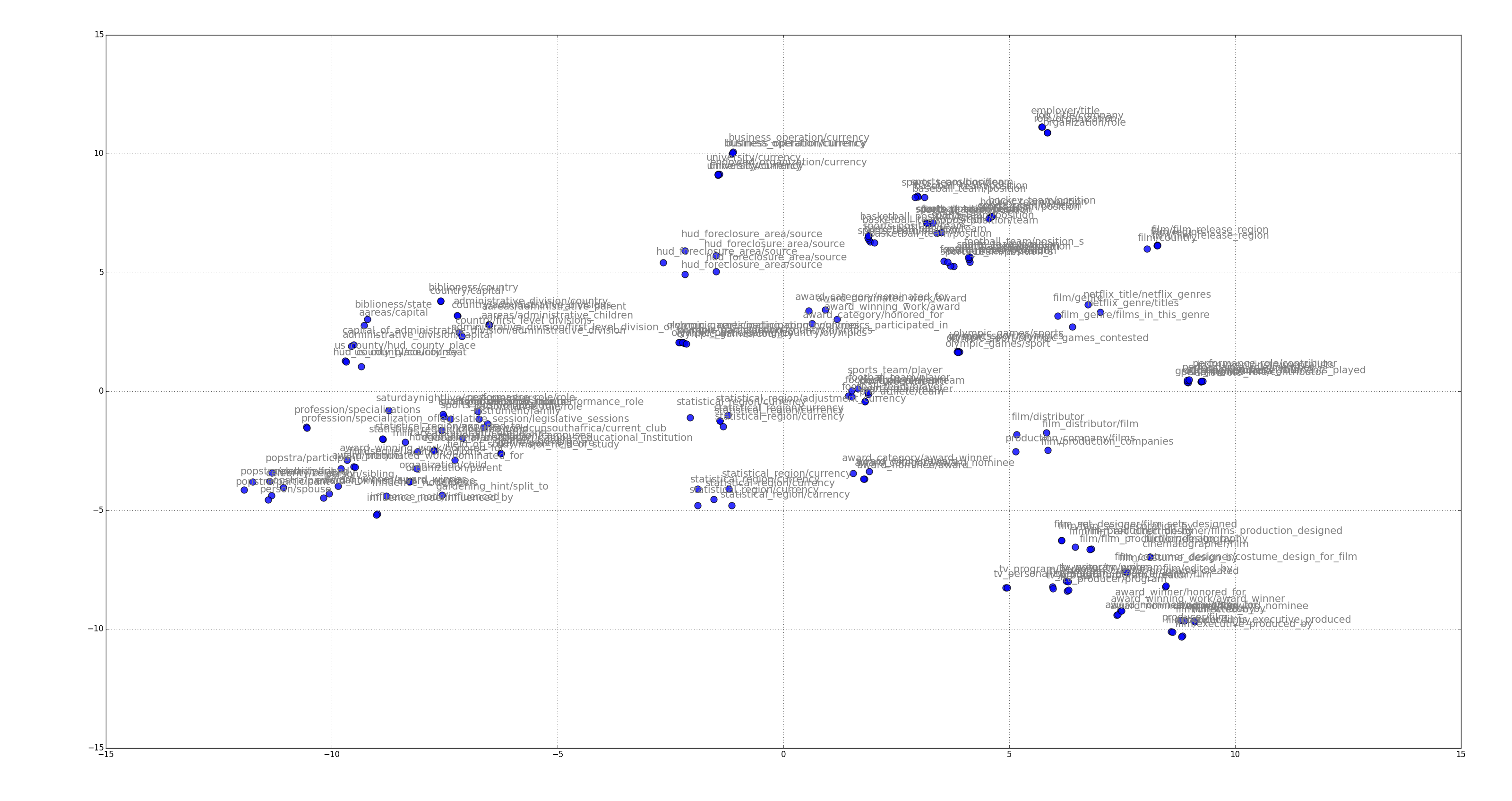}
\caption{Relation embeddings ({\sc DistMult})}
\label{fig:distmult}
\end{figure}

Figure~\ref{fig:distadd} and~\ref{fig:distmult} illustrate the relation embeddings learned by {\sc DistMult} and {\sc DistAdd} using t-SNE. We selected $189$ relations in the FB15k-401 dataset. The embeddings learned by {\sc DistMult} nicely reflect the clustering structures among these relations (e.g. /film/release\_region is closed to /film/country); whereas the embeddings learned by {\sc DistAdd} present structure that is harder to interpret. 


Table~\ref{rel_example} shows some concrete examples: top $3$ nearest neighbors in the relation space learned by {\sc DistMult} and {\sc DistAdd} along with the distance values (Frobenius distance between two relation matrices or Euclidean distance between two relation vectors). We can see that the nearest neighbors found by {\sc DistMult} are much more meaningful. {\sc DistAdd} tends to retrieve irrelevant relations which take in completely different types of arguments.
\begin{table*}[bth]
\begin{center}
\fontsize{8}{7.2}\selectfont
\begin{tabular}{|c|c|c|}
\hline
& {\sc DistMult} & {\sc DistAdd} \\
\hline
/film\_distributor/film  & \parbox{5cm}{/film/distributor (2.0)\\/production\_company/films (3.4)\\ /film/production\_companies (3.4)} & \parbox{5cm}{/production\_company/films (2.6)\\ /award\_nominee/nominated\_for (2.7)\\/award\_winner/honored\_for (2.9)}\\ 
\hline
/film/film\_set\_decoration\_by & \parbox{5cm}{/film\_set\_designer/film\_sets\_designed (2.5) \\ /film/film\_art\_direction\_by (6.8) \\ /film/film\_production\_design\_by (9.6)} & \parbox{5cm}{/award\_nominated\_work/award\_nominee (2.7)\\ /film/film\_art\_direction\_by (2.7)\\ /award\_winning\_work/award\_winner (2.8)}\\
\hline
/organization/leadership/role & \parbox{5cm}{/leadership/organization (2.3)\\ /job\_title/company (12.5)\\ /business/employment\_tenure/title (13.0) } & \parbox{5cm}{/organization/currency (3.0) \\ /location/containedby	 (3.0) \\ /university/currency (3.0) }\\
\hline
/person/place\_of\_birth & \parbox{5cm}{/location/people\_born\_here (1.7)\\ /person/places\_lived (8.0) \\ /people/marriage/location\_of\_ceremony (14.0) } & \parbox{5cm}{/us\_county/county\_seat (2.6)\\ /administrative\_division/capital (2.7) \\ /educational\_institution/campuses (2.8)}\\
\hline
\end{tabular}
\caption{\label{rel_example}Examples of nearest neighbors of relation embeddings}
\end{center}
\end{table*}

\end{document}